\documentclass{article}
\usepackage{spconf,amsmath,graphicx,url}
\usepackage{enumitem,booktabs,multirow,hyperref,setspace}
\usepackage[caption=false]{subfig}

\usepackage[
backend=biber,
style=ieee,
maxbibnames=3,
maxcitenames=3,
doi=false,isbn=false,url=false,eprint=false
]{biblatex} 

\defbibheading{bibliography}[\refname]{}

\DeclareSourcemap{
	\maps[datatype=bibtex, overwrite=true]{
		\map{
			\step[fieldsource=booktitle,
			match=\regexp{.*Interspeech.*},
			replace={Proc. Interspeech}]
			\step[fieldsource=journal,
			match=\regexp{.*INTERSPEECH.*},
			replace={Proc. Interspeech}]
			\step[fieldsource=booktitle,
			match=\regexp{.*ICASSP.*},
			replace={Proc. ICASSP}]
			\step[fieldsource=booktitle,
			match=\regexp{.*icassp_inpress.*},
			replace={Proc. ICASSP (in press)}]
			\step[fieldsource=booktitle,
			match=\regexp{.*International.*Conference.*on.*Acoustics,.*Speech.*and.*Signal.*Processing.*},
			replace={Proc. ICASSP}]
			\step[fieldsource=booktitle,
			match=\regexp{.*International.*Conference.*on.*Learning.*Representations.*},
			replace={Proc. ICLR}]
			\step[fieldsource=booktitle,
			match=\regexp{.*International.*Conference.*on.*Machine.*Learning.*},
			replace={Proc. ICML}]
			\step[fieldsource=booktitle,
			match=\regexp{.*International.*conference.*on.*machine.*learning.*},
			replace={Proc. ICML}]
			\step[fieldsource=booktitle,
			match=\regexp{.*Automatic.*Speech.*Recognition.*and.*Understanding.*},
			replace={Proc. ASRU}]
			\step[fieldsource=booktitle,
			match=\regexp{.*Spoken.*Language.*Technology.*},
			replace={Proc. SLT}]
			\step[fieldsource=booktitle,
			match=\regexp{.*Speech.*Synthesis.*Workshop.*},
			replace={Proc. SSW}]
			\step[fieldsource=booktitle,
			match=\regexp{.*workshop.*on.*speech.*synthesis.*},
			replace={Proc. SSW}]
			\step[fieldsource=booktitle,
			match=\regexp{.*Advances.*in.*neural.*information.*processing.*},
			replace={Proc. NeurIPS}]
			\step[fieldsource=booktitle,
			match=\regexp{.*Advances.*in.*Neural.*Information.*Processing.*},
			replace={Proc. NeurIPS}]
			\step[fieldsource=booktitle,
			match=\regexp{.*Workshop.*on.* Applications.* of.* Signal.*Processing.*to.*Audio.*and.*Acoustics.*},
			replace={Proc. WASPAA}]
			\step[fieldsource=publisher,
			match=\regexp{.+},
			replace={{}}]
			\step[fieldsource=month,
			match=\regexp{.+},
			replace={{}}]
			\step[fieldsource=location,
			match=\regexp{.+},
			replace={{}}]
			\step[fieldsource=address,
			match=\regexp{.+},
			replace={{}}]
			\step[fieldsource=organization,
			match=\regexp{.+},
			replace={{}}]
		}
	}
}

\title{ESPnet2-TTS: Extending the Edge of TTS Research}

\makeatletter
\def\@name{
  \emph{Tomoki Hayashi}$^{1,2}$,
  \emph{Ryuichi Yamamoto}$^{3}$,
  \emph{Takenori Yoshimura}$^{4}$,
  \emph{Peter Wu}$^{5}$,
  \emph{Jiatong Shi}$^{5}$, \\
  \emph{Takaaki Saeki}$^{6}$,
  \emph{Yooncheol Ju}$^{7}$,
  \emph{Yusuke Yasuda}$^{2}$,
  \emph{Shinnosuke Takamichi}$^{6}$,
  \emph{Shinji Watanabe}$^{5}$
}
\makeatother
\address{
  $^1$Human Dataware Lab. Co., Ltd.,
  $^2$Nagoya University,
  $^3$LINE Corp.,
  $^4$Nagoya Institute of Technology, \\
  $^5$Carnegie Mellon University,
  $^6$The University of Tokyo,
  $^7$AIRS Company, Hyundai Motor Group
}

\begin{document}
\ninept
\maketitle
\begin{abstract}
This paper describes ESPnet2-TTS, an end-to-end text-to-speech (E2E-TTS) toolkit. ESPnet2-TTS extends our earlier version, ESPnet-TTS, by adding many new features, including: on-the-fly flexible pre-processing, joint training with neural vocoders, and state-of-the-art TTS models with extensions like full-band E2E text-to-waveform modeling, which simplify the training pipeline and further enhance TTS performance.
The unified design of our recipes enables users to quickly reproduce state-of-the-art E2E-TTS results.
We also provide many pre-trained models in a unified Python interface for inference, offering a quick means for users to generate baseline samples and build demos.
Experimental evaluations with English and Japanese corpora demonstrate that our provided models synthesize utterances comparable to ground-truth ones, achieving state-of-the-art TTS performance. 
The toolkit is available online at \url{https://github.com/espnet/espnet}.
\end{abstract}
%

\section{Introduction}
\label{sec:intro}

Thanks to the improvements of deep learning techniques, end-to-end text-to-speech (E2E-TTS) models have replaced the conventional statistical parametric speech synthesis (SPSS) systems based on hidden Markov models (HMMs)~\cite{tokuda2013speech} and deep neural networks (DNNs)~\cite{zen2014deep}.
E2E-TTS models can alleviate complex text pre-processing and learn alignments between input and output sequences in a data-driven manner, achieving high-fidelity speech comparable to professional recordings~\cite{wang2017tacotron,shen2017tacotron2,li2018transformer}. 
Recent E2E-TTS works have also introduced a range of desirable TTS features.
For example, non-autoregressive (NAR) architectures enable us to generate speech faster than real-time~\cite{ren2019fastspeech,peng2020non,ren2020fastspeech}.
Additionally, introducing learnable global or fine-grained embeddings allows us to control speaker characteristics~\cite{jia2018transfer,chen2020multispeech} or speaking styles~\cite{wang2018style,sun2020fully}.

To help researchers accelerate the development of TTS techniques and stay up to date with the state-of-the-art models, we first developed an open-source E2E-TTS toolkit called ESPnet-TTS~\cite{hayashi2020espnet}, which followed the great success of the speech recognition toolkits Kaldi~\cite{Povey_ASRU2011_2011} and ESPnet-ASR~\cite{watanabe2018espnet}.
This toolkit provided Chainer~\cite{tokui2019chainer} and PyTorch~\cite{paszke2019pytorch}-based neural network libraries and highly reproducible recipes.
ESPnet-TTS also contributed to many research projects and development platforms for new applications like voice conversion~\cite{zhao2020voice,huang2019voice}.
However, since the toolkit required a fair amount of offline processing, such as feature extraction and text frontend processing, there existed room for improvement in terms of scalability, flexibility, and portability.

In this paper, we introduce our second generation TTS toolkit named ESPnet2-TTS.
Compared to our previous toolkit~\cite{hayashi2020espnet}, it includes many new features such as on-the-fly pre-processing, the Model Zoo, and state-of-the-art models with our own extensions.
The contributions of this paper are summarized as follows:
\begin{itemize}[leftmargin=3mm,itemsep=0mm]
    \item We introduce the ESPnet2-TTS toolkit, containing new features like a unified task design, flexible on-the-fly pre-processing, and a simple Python interface to quickly use many pre-trained models in the Model Zoo.
    
    \item We introduce our provided state-of-the-art models, including E2E text-to-waveform (T2W) models and joint training with neural vocoders, which simplify the training pipeline and enhance TTS performance.
    Furthermore, we provide extensions such as the Conformer architecture~\cite{gulati2020conformer}, full-band waveform modeling, and zero-shot adaptation with pre-trained speaker embeddings.
    
    \item We conduct experimental evaluations with English and Japanese corpora, and investigate the performance in various settings, including single speaker, multi-speaker, and adaptation. Evaluation results demonstrate that the model can achieve state-of-the-art performance comparable with the ground-truth.
\end{itemize}

\section{Related works}
We observe a number of E2E-TTS toolkits whose functions are similar to ours, e.g., TensorFlow-TTS\footnote{\url{https://github.com/TensorSpeech/TensorFlowTTS}}, coqui-ai TTS\footnote{\url{https://github.com/coqui-ai/TTS}}, OpenSeq2Seq~\cite{kuchaiev2018openseq2seq}, NeMo~\cite{kuchaiev2019nemo}, and Fairseq S$^2$~\cite{wang2021fairseq}.
While Tensorflow-TTS and coqui-ai TTS focus on only TTS, the others support several speech processing tasks such as automatic speech recognition (ASR), machine translation (MT), and speech translation (ST).
Compared with these toolkits, ESPnet2-TTS provides a number of reproducible recipes, including various languages and scenarios (e.g., single speaker, multi-speaker, and adaptation).
This helps researchers quickly test state-of-the-art TTS models in major and minor languages, giving lots of insights for accelerating the development of new TTS research ideas.
ESPnet2-TTS also supports various state-of-the-art TTS models, including E2E-T2W models and joint training with neural vocoders.
Moreover, various of our own extensions are available, such as the Conformer architecture~\cite{guo2021recent}, full-band waveform generation, and zero-shot adaptation with pre-trained speaker embeddings.
Furthermore, ESPnet2 supports various speech processing tasks, including ASR, speech enhancement, speech diarization, and self-supervised learning, enabling users to build such models through the same interface as TTS.

\section{Features of ESPnet2-TTS}

\subsection{ESPnet2}
ESPnet2 is a new framework for network training shared between various speech processing tasks.
To enhance the scalability, flexibility and portability of our previous toolkit~\cite{hayashi2020espnet}, ESPnet2 introduces multiple new features, as described below.
\begin{description}[style=unboxed,leftmargin=0mm,itemsep=0mm]
    \item[Unified task design] Inspired by FairSeq~\cite{ott2019fairseq}, ESPnet2 has a unified task interface to define new speech processing tasks quickly.
    The task interface abstracts the data loader and the training iterations to handle complex task configurations.
    Specifically, this makes it easy to build a complicated model requiring various inputs and performing multi-node multi-GPU training for a more extensive network.

    \item[On-the-fly] ESPnet2 can perform pre-processing in an on-the-fly manner during training.
    Pre-processing features include text cleaning, grapheme-to-phoneme (G2P) conversion, acoustic feature extraction, and data augmentation, and each pre-processing step can be easily customized (e.g., adding a new G2P function). 
    The on-the-fly pre-processing allows launching multiple training jobs without creating sizeable temporary files for each setting and makes it easy to deploy the model with a simple interface.
    
    \item[Model Zoo] Inspired by Asteroid~\cite{Pariente2020Asteroid} and HuggingFace transformers~\cite{wolf-etal-2020-transformers}, ESPnet2 has a Model Zoo that provides quick access to pre-trained models so that users can utilize pre-trained models with just a few lines of code.
\end{description}

\subsection{Recipes}
Based on a Kaldi-style file structure, our recipes are based on shell scripts that perform all the steps required to reproduce the results.
To reduce maintenance costs and accelerate the creation of new recipes, ESPnet2 provides a template shared across all recipes.
Namely, users can create a recipe by just modifying the script for the data preparation stage since the code for the other stages is the same for all recipes.
ESPnet2-TTS provides 20+ recipes that cover 10+ languages, including single-speaker, multi-speaker, multi-language, and adaptation use cases.

\subsection{Models}
ESPnet2-TTS supports not only standard text-to-mel (T2M) and mel-to-waveform (M2W) models, but also state-of-the-art models such as E2E-T2W with our own extensions.

\begin{description}[style=unboxed,leftmargin=0mm,itemsep=0mm]
\item[T2M models]
T2M models generate mel spectrograms from input text representations (e.g., phonemes).
ESPnet2-TTS supports two autoregressive (AR) T2M models, Tacotron~2~\cite{shen2017tacotron2} and Transformer-TTS~\cite{li2018transformer}, and two non-autoregressive (NAR) T2M models, FastSpeech~\cite{ren2019fastspeech} and FastSpeech~2~\cite{ren2020fastspeech} extended with techniques from FastPitch~\cite{lancucki2021fastpitch}.
The extended FastSpeech~2 predicts token-averaged energy and pitch sequences instead of raw pitch and energy sequences.
Furthermore, we extend these NAR models with the Conformer~\cite{gulati2020conformer} architecture, dubbed Conformer-FastSpeech and Conformer-FastSpeech~2~\cite{guo2021recent}.
For AR models, we use guided attention loss to help to learn diagonal attentions ~\cite{tachibana2018efficiently}.
For NAR models, we use the duration of each input token calculated from the attention weights of AR models~\cite{ren2019fastspeech}.
Although this requires building the AR models before training NAR models, we do not need to prepare the external aligner and use any input representations.
Furthermore, we provide multi-speaker extensions using one-hot speaker embeddings, pre-trained X-vectors~\cite{snyder2018x}, and global style tokens (GSTs)~\cite{wang2018style} for all T2M models.

\item[M2W models]
M2W models, also known as (neural) vocoders, generate waveforms from mel spectrograms.
For M2W modelling, we support the Griffin-Lim vocoder and NAR neural vocoders using generative adversarial networks (GANs).
The latter includes Parallel WaveGAN~\cite{yamamoto2020parallel}, MelGAN~\cite{kumar2019melgan}, StyleMelGAN~\cite{mustafa2021stylemelgan}, HiFi-GAN~\cite{kong2020hifi}, and their multi-band extensions~\cite{yang2021multi}.
Since GAN-based models consist of generators and discriminators, we provide support for training arbitrary combinations of generators and discriminators.

\item[Joint-T2W models]
ESPnet2-TTS provides a function to jointly train T2M and M2W models, which we refer to as the joint text-to-waveform (Joint-T2W) model.
For Joint-T2W models, we adopt a GAN-based training strategy using random windowed discriminators (RWDs)~\cite{binkowski2019high}. 
The Joint-T2W models are optimized to minimize the sum of the T2M model loss, calculated using whole sequence, and the M2W model loss, calculated from randomly windowed subsequences.
The use of subsequences for the M2W loss alleviates the issue of huge memory consumption due to the length of the waveforms~\cite{chen2021wavegrad}.
We can combine an arbitrary T2M model with an arbitrary neural M2W one. 
Also, we can use this joint training not only for fine-tuning but also for training from scratch.
This joint training can simplify the training scheme of TTS models and enhance their performance.

\item[E2E-T2W models]
E2E-T2W models perform TTS in a truly end-to-end fashion, generating waveforms from input text representations.
For E2E-T2W models, we support VITS~\cite{kim2021conditional}, which utilizes a conditional variational autoencoder (CVAE) with normalizing flows and GAN-based optimizations. 
We further extend VITS to make it possible to use the Conformer architecture for the text encoder, generate a full-band waveform (44.1 kHz), and perform zero-shot adaptation with pre-trained X-vectors.

\end{description}

\subsection{Evaluation}
To evaluate TTS model performance, ESPnet2-TTS provides three objective evaluation metrics: Mel-cepstral distortion (MCD), log-$F_0$ root mean square error ($F_0$ RMSE), and character error rate (CER) with pre-trained ESPnet2-ASR models.
MCD and $F_0$ RMSE reflect speaker, prosody, and phonetic content similarities, and CER can reflect the intelligibility of generated speech.
For MCD and $F_0$ RMSE, we apply dynamic time-warping (DTW)~\cite{salvador2007toward} to match the length difference between ground-truth speech and generated speech.
While these objective metrics can estimate the quality of synthesized speech, it is still difficult to fully determine human perceptual quality from these values, especially with high-fidelity generated speech.
Hence, we also provide instructions for subjective evaluations with mean opinion scores (MOS) based on webMUSHRA~\cite{schoeffler2018webmushra}. 
These instructions help researchers quickly launch a web-based subjective evaluation system and collect subjects through crowdsourcing services like Amazon Mechanical Turk.

\section{Experimental evaluation}
To study the performance of our TTS models, we conducted experimental evaluations with English and Japanese corpora.
The objective evaluation metrics were MCD, $F_0$ RMSE, and CER.
The subjective evaluation metric was MOS for naturalness with a 5-point scale: 5 for excellent, 4 for good, 3 for fair, 2 for poor, and 1 for bad.

We made the pretrained weights and configurations of all models public for reproducibility.
Audio samples are available online\footnote{\url{https://espnet.github.io/icassp2022-tts}}.

\subsection{English single speaker}
First, we evaluated the performance of English single-speaker models with the LJSpeech dataset~\cite{ito2017ljspeech}, which consists of 24 hours of speech recorded with 16 bits and a 22.05 kHz sampling rate.
We followed the recipe in \url{egs2/ljspeech/tts1}, using 12,600 utterances for training, 250 for validation, and 250 for evaluation.
Also, we used g2p-en\footnote{\url{https://github.com/Kyubyong/g2p}} without word separators as the G2P function.
We compared the following models:
\begin{description}[style=unboxed,leftmargin=2mm,itemsep=0mm,topsep=1pt]
\item[Transformer~\cite{hayashi2020espnet}] Our previous best model in \cite{hayashi2020espnet}, which consists of Transformer-TTS and the mixture of logistics WaveNet vocoder~\cite{shen2017tacotron2}. To enhance the perceptual quality, the noise shaping technique was used to mask noise in the high frequency band~\cite{tachibana2018invetigation}.
\item[CFS2] Conformer-FastSpeech2 (CFS2) + HiFi-GAN. Each of these parts was trained separately. The duration of each token was calculated from a Tacotron~2 teacher model.
\item[CFS2 (+ft)] Same as the above combination, but HiFi-GAN was fine-tuned with ground-truth aligned outputs generated by CFS2.
\item[CFS2 (+joint-ft)] Same as the above combination, but the two networks were jointly fine-tuned.
\item[CFS2 (+joint-tr)] Same as the above combination, but the two networks were jointly trained from scratch.
\item[VITS] The end-to-end text-to-waveform model VITS. We used 0.333 of the scaling factor for the standard deviation of the stochastic duration predictor and the prior distribution.
\end{description}
For the CER calculation, we used an ESPnet2-ASR pre-trained model\footnote{\url{https://zenodo.org/record/4030677}} trained on the LibriSpeech dataset~\cite{panayotov2015librispeech}.
Since the ASR model assumed that its speech inputs had sample rates of 16~kHz, we downsampled audio to 16~kHz before using the ASR model.
For subjective evaluation, we randomly selected 60 utterances from the evaluation set and had 37 English speakers as listeners.

\begin{table}[t!]\centering
\caption{\it Results on LJSpeech corpus, where ``STD'' represents standard deviation and ``CI'' represents 95 \% confidence intervals.}\label{tb:ljspeech_results}
\vspace{1pt}
\scalebox{0.78}{
\begin{tabular}{lcccc}\toprule
Method &MCD ± STD &$F_0$ RMSE ± STD &CER &MOS ± CI \\
\midrule
GT &N/A &N/A &1.1 &4.15 ± 0.08 \\
Transformer~\cite{hayashi2020espnet} &6.97 ± 0.79 &0.252 ± 0.042 &2.7 &3.86 ± 0.08 \\
\midrule
CFS2 &\textbf{6.47 ± 0.58} &\textbf{0.214 ± 0.031} &\textbf{1.2} &3.53 ± 0.09 \\
CFS2 (+ft) &6.51 ± 0.58 &0.217 ± 0.032 &\textbf{1.2} &4.00 ± 0.07 \\
CFS2 (+joint-ft) &6.73 ± 0.62 &0.221 ± 0.032 &1.5 &\textbf{4.03 ± 0.07} \\
CFS2 (+joint-tr) &6.80 ± 0.54 &0.218 ± 0.032 &1.5 &3.92 ± 0.08 \\
VITS &6.84 ± 0.65 &0.232 ± 0.033 &2.7 &3.88 ± 0.08 \\
\bottomrule
\end{tabular}
}
\vspace{-5mm}
\end{table}

\begin{figure}[t!]
  \centering
  \subfloat[\textit{Generated spectrogram by VITS.}]{\includegraphics[width=0.48\columnwidth]{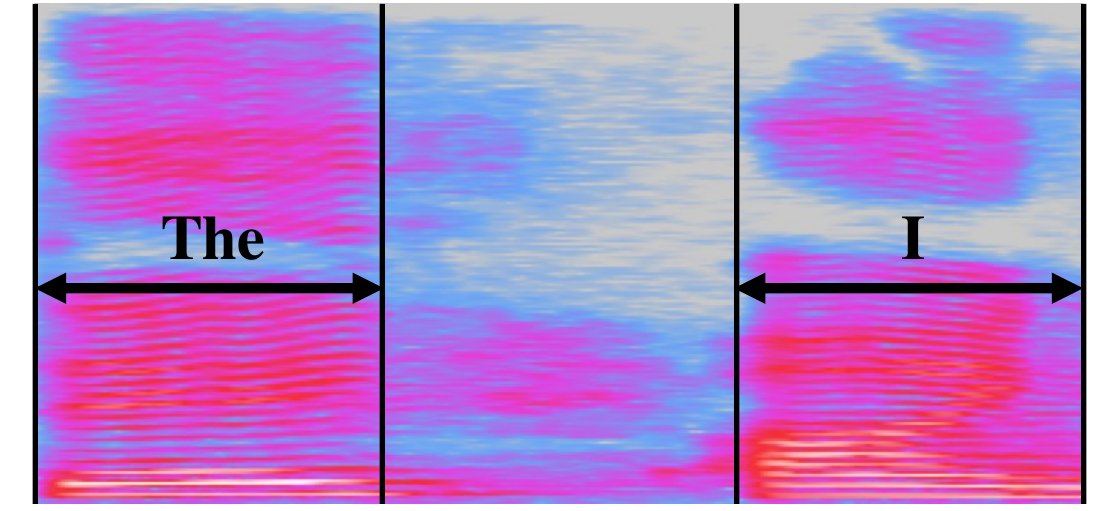}}\hspace{2mm}
  \subfloat[\textit{Generated spectrogram by CFS2.}]{\includegraphics[width=0.48\columnwidth]{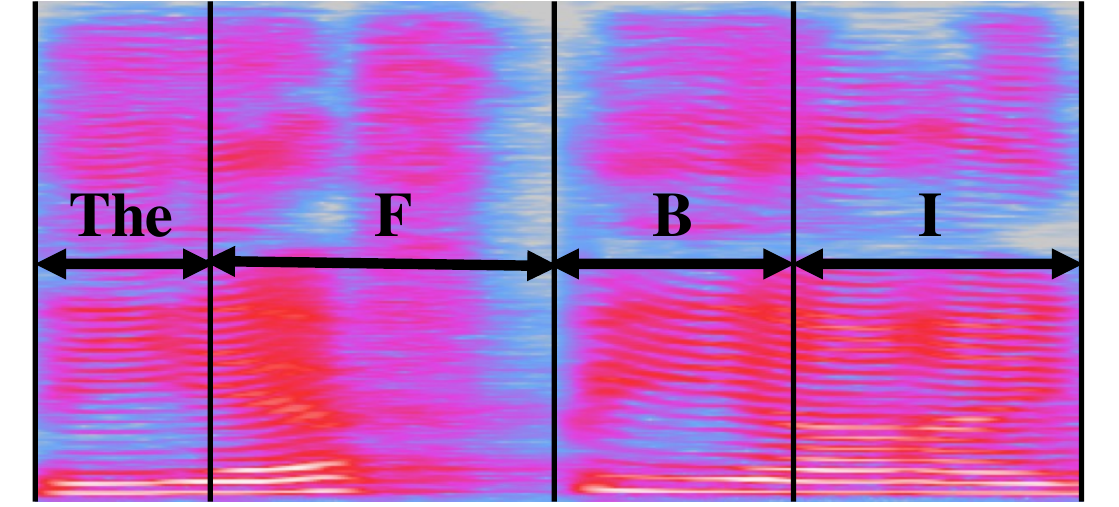}}
  \vspace{-2mm}
  \caption{\it Generated example using the wrong G2P result. The input text is ``The FBI'' and g2p-en output is ``DH AH0 B AY1'', which lacked phonemes corresponding to the character ``F’'.}
  \label{fig:g2p_recover_example}
  \vspace{-5mm}
\end{figure}

The experimental evaluation results are shown in Table~\ref{tb:ljspeech_results}.
Our new best model outperformed our previous study's best model, achieving comparable intelligibility and naturalness with the ground-truth.
Moreover, since CFS2 and VITS consist of NAR architectures, they can perform inference much faster than our previous model, which requires slow AR computations.
With joint training, CFS2 (+joint-tr) solves the mismatch of acoustic features between training and evaluation and significantly outperforms CFS2 without joint training in terms of naturalness.
The fine-tuning of HiFi-GAN worked well, and joint fine-tuning helped further reduce the metallic noise appearing in fricative consonant sounds such as /s/ and /z/, improving the naturalness slightly more.
With VITS, there is a slight gap between the ground truth and the synthesized speech for both intelligibility and naturalness.
We found that one of the reasons for this gap was the mispronunciation of words that included G2P errors.
Fig.~\ref{fig:g2p_recover_example} shows an example of the generated spectrogram when the input contains G2P errors.
Interestingly, CFS2 can generate correct pronunciations even with the wrong G2P results; however, VITS misses some characters.
This implies that CFS2 can recover from incorrect G2P results through training thanks to the soft alignment derived from the teacher AR model.
In contrast, VITS is more sensitive to G2P errors because of the hard monotonic alignment.
To check this hypothesis, we investigated the performance when using a slower but more accurate G2P function (espeak-ng\footnote{\url{https://github.com/espeak-ng/espeak-ng}}) and the results are shown in Table~\ref{tb:ljspeech_results_2}.
From these results, using a better G2P function improved the intelligibility of VITS while CFS2 had a similar intelligibility, reinforcing our hypothesis. 

\begin{table}[t!]\centering
\caption{\it Results with different G2P functions on LJSpeech corpus, where ``STD'' represents standard deviation.}\label{tb:ljspeech_results_2}
\vspace{1pt}
\scalebox{0.79}{
\begin{tabular}{lccc}\toprule
Method &MCD ± STD &$F_0$ RMSE ± STD &CER \\
\midrule
CFS2 (g2p-en) & \textbf{6.47 ± 0.58} & \textbf{0.214 ± 0.031} &1.2 \\
CFS2 (espeak-ng) & 6.51 ± 0.62 & 0.216 ± 0.033 & \textbf{1.1} \\
VITS (g2p-en) &6.84 ± 0.65 &0.232 ± 0.033 &2.7 \\
VITS (espeak-ng) &6.70 ± 0.58 &0.228 ± 0.037 &1.6 \\
\bottomrule
\end{tabular}
}
\vspace{-5mm}
\end{table}

\subsection{English multi-speaker}
Next, we investigated the performance of multi-speaker VITS with two types of speaker embeddings using the VCTK dataset~\cite{yamagishi2019cstr}.
We followed the recipe in \url{egs2/vctk/tts1} for pre-processing and training.
To evaluate the models in seen and unseen speaker conditions, we randomly selected four speakers (two males and two females) as the seen speaker evaluation set and another four as the unseen speaker one.
The number of evaluation utterances per speaker was set to 10 so that the total number of utterances was 40 for each condition.
We downsampled audio to 22.05 kHz and used espeak-ng as the G2P function, which includes phoneme and stress symbols.
We compared the following models:
\begin{description}[style=unboxed,leftmargin=2mm,itemsep=0mm,topsep=1pt]
    \item[SID-VITS] VITS with one-hot speaker ID (SID) embeddings. Since this model cannot deal with unknown speakers, we trained it with all of the speakers.
    \item[X-VITS (Avg.)] VITS with pre-trained X-vectors instead of one-hot speaker ID embeddings.
    This model was trained with all speakers except for the evaluation ones in the unseen speaker condition.
    For inference, we used X-vectors averaged over all the utterances of the target speaker except for the evaluation utterances. 
    \item[X-VITS (Ran.)] The same as the above model except it used X-vectors extracted from a single utterance of the target speaker. This datapoint was randomly selected from all utterances of the speaker excluding the evaluation utterances.
\end{description}
For CER computation, we used the same ASR model as the one described in Section 4.1.
In total, 27 English speakers participated in the subjective evaluation listening test.

Tables~\ref{tb:vctk_results_closed} and \ref{tb:vctk_results_open} show the evaluation results for seen and unseen speaker conditions, respectively.
In the former, both VITS models achieved naturalness comparable to the ground-truth, and 
using X-vectors extracted with more utterances improved speaker similarity since it yielded lower MCD and $F_0$ RMSE.
The results from the latter show that even for unknown speakers, X-VITS can generate natural speech and improve speaker similarity by using more reference utterances from the target speaker.

\begin{table}[t!]\centering
\caption{\it Results of the seen speaker condition on VCTK corpus, where ``STD'' represents standard deviation and ``CI'' represents 95 \% confidence intervals.}
\label{tb:vctk_results_closed}
\vspace{1pt}
\scalebox{0.79}{
\begin{tabular}{lcccc}\toprule
Method &MCD ± STD &$F_0$ RMSE ± STD &CER &MOS ± CI \\
\midrule
GT &N/A &N/A &3.9 &4.03 ± 0.08 \\
\midrule
SID-VITS &\textbf{6.30 ± 0.82} &\textbf{0.242 ± 0.110} &\textbf{5.5} &\textbf{4.00 ± 0.08}\\
X-VITS (Avg.) &6.35 ± 0.91 &0.257 ± 0.119 &6.3 &3.99 ± 0.08 \\
X-VITS (Ran.) &6.95 ± 1.09 &0.289 ± 0.123 &6.9 &3.96 ± 0.08 \\
\bottomrule
\end{tabular}
}
\vspace{-5mm}
\end{table}

\begin{table}[!t]\centering
\caption{\it Results of the unseen speaker condition on VCTK corpus, where ``STD'' represents standard deviation and ``CI'' represents 95 \% confidence interval. Note that SID-VITS used all speakers.}
\label{tb:vctk_results_open}
\vspace{1pt}
\scalebox{0.79}{
\begin{tabular}{lcccc}\toprule
Method &MCD ± STD &$F_0$ RMSE ± STD &CER &MOS ± CI \\
\midrule
GT &N/A &N/A &2.8 & 4.04 ± 0.08\\
\midrule
SID-VITS &\textbf{6.91 ± 1.48} &0.275 ± 0.101 &7.0 &3.92 ± 0.08\\
X-VITS (Avg.) &7.41 ± 0.83 &\textbf{0.274 ± 0.110} &\textbf{4.9} &\textbf{4.04 ± 0.08} \\
X-VITS (Ran.) &7.89 ± 1.14 &0.289 ± 0.128 &8.0 &3.97 ± 0.08 \\
\bottomrule
\end{tabular}
}
\vspace{-4mm}
\end{table}

\subsection{Japanese single speaker}
Next, we evaluated the performance of Japanese single-speaker models with the JSUT corpus~\cite{sonobe2017jsut}, which consists of 10 hours of speech recorded with 16 bits and a 48 kHz sampling rate.
We followed the recipe in \url{egs2/jsut/tts1}, using 7,196 utterances for training, 250 for validation, and 250 for evaluation. 
We used the G2P function based on Open~JTalk enhanced with prosody symbols~\cite{kiyoshi2021prosodic} for all models.
We compared the following architectures:
\begin{description}[style=unboxed,leftmargin=2mm,itemsep=0mm,topsep=1pt]
\item[Tacotron~2] Tacotron~2 + HiFi-GAN. Each model was separately trained with a sampling rate of 24 kHz.
\item[Transformer] Transformer-TTS + HiFi-GAN. Each model was separately trained with a sampling rate of 24 kHz.
\item[CFS2] Conformer-FastSpeech2 + HiFi-GAN. Each model was separately trained with a sampling rate of 24 kHz.
\item[CFS2 (+ft)] Same as the above model, but HiFi-GAN was fine-tuned with ground-truth aligned mel spectrograms.
\item[VITS] VITS trained with a sampling rate of 22.05 kHz.
\item[FB-VITS] Full-band VITS trained with a sampling rate of 44.1 kHz.
\end{description}
To calculate CER, we used an ESPnet2-ASR pre-trained model\footnote{\url{https://zenodo.org/record/4037458}} trained on the Corpus of Spontaneous Japanese (CSJ)~\cite{maekawa2003csj}.
For subjective evaluation, we randomly selected 60 utterances from the evaluation set.
We also downsampled from a frequency of 24 kHz to 22.05 kHz in order to match model sampling rate assumptions.
In total, 37 Japanese native speakers evaluated these models.

\begin{table}[t!]
\centering
\caption{\it Results on JSUT corpus, where ``STD'' represents standard deviation, ``CI'' represents 95 \% confidence interval, and ``$*$'' represents different analysis conditions due to the sampling rate.}\label{tb:jsut_results}
\vspace{1pt}
\scalebox{0.79}{
\begin{tabular}{lcccc}\toprule
Method &MCD ± STD & $F_0$ RMSE ± STD &CER &MOS ± CI\\
\midrule
GT (22k) &N/A &N/A &6.0 &4.02 ± 0.08 \\
GT (44k) &N/A &N/A &6.0 &4.02 ± 0.08 \\
\midrule
Tacotron~2 &6.62 ± 0.60 &0.177 ± 0.036 &7.0 &3.54 ± 0.09 \\
Transformer &6.58 ± 0.62 &0.179 ± 0.038 &7.2 &3.58 ± 0.09 \\
CFS2 &\textbf{6.26 ± 0.57} &0.158 ± 0.034 &\textbf{6.3} &3.79 ± 0.08 \\
CFS2 (+ft) &6.34 ± 0.57 &0.158 ± 0.032 &\textbf{6.3} &3.86 ± 0.08 \\
VITS &6.37 ± 0.55 &\textbf{0.157 ± 0.033} &6.9 &\textbf{4.00 ± 0.08} \\
\midrule
FB-VITS & 6.24 ± 0.38$^*$ &0.158 ± 0.031$^*$ &7.2 &\textbf{4.00 ± 0.08} \\
\bottomrule
\end{tabular}
}
\vspace{-5mm}
\end{table}

Table~\ref{tb:jsut_results} contains the evaluation results.
While CFS2 achieves the best intelligibility, VITS produces the best naturalness, comparable to the ground-truth.
Interestingly, adding the full-band extension did not improve naturalness.
However, since we observed that the listening equipment might affect the evaluation, we recommend listening to our samples and comparing the difference between 22.05 kHz and 44.1 kHz ones.

\subsection{Japanese single speaker adaptation}
Finally, we investigate the performance of VITS in speaker adaptation with a small amount of training data from the JVS corpus~\cite{takamichi2019jvs}.
We followed the recipe in \url{egs2/jvs/tts1} and selected four speakers (two males and two females), using 100 utterances for adaptation and 30 for evaluation.
For the pre-trained model, we used VITS trained on JSUT with 22.05 kHz audio.
For the CER calculation, we used the same model as the one from Section 4.3.
In total, 28 Japanese native speakers participated in the subjective evaluation.

\begin{table}[!t]\centering
\caption{\it Evaluation results of VITS adaptation with JVS corpus, where ``STD'' represents standard deviation and ``CI'' represents 95 \% confidence interval. The ``jvs001'' and ``jvs054'' are male speakers, and the rest are female ones.}\label{tb:jvs_results}
\vspace{1pt}
\scalebox{0.79}{
\begin{tabular}{lcccc}\toprule
Method &MCD ± STD & $F_0$ RMSE ± STD &CER &MOS ± CI\\
\midrule
GT (jvs001) &N/A &N/A &7.2 &4.73 ± 0.06 \\
VITS (jvs001) &5.75 ± 0.65 &0.210 ± 0.055 &7.5 &3.37 ± 0.13 \\
\midrule
GT (jvs054) &N/A &N/A &5.9 &4.49 ± 0.09 \\
VITS (jvs054) &5.30 ± 0.40 &0.257 ± 0.042 &8.8 &3.44 ± 0.13 \\
\midrule
GT (jvs010) &N/A &N/A &4.3 &3.91 ± 0.12 \\
VITS (jvs010) &5.89 ± 0.39 &0.144 ± 0.026 &3.8 &3.31 ± 0.13 \\
\midrule
GT (jvs092) &N/A &N/A &7.3 &4.05 ± 0.11 \\
VITS (jvs092) &5.49 ± 0.50 &0.140 ± 0.034 &6.7 &3.45 ± 0.11 \\
\bottomrule
\end{tabular}
}
\vspace{-4mm}
\end{table}

Table~\ref{tb:jvs_results} shows the evaluation results. 
We find that adaptation with a small amount of training data also works for VITS, achieving comparable intelligibility and reasonable naturalness.
Though there is a gap between the ground-truth in terms of naturalness, thanks to the E2E-T2W architecture, we do not need to perform adaptation with the T2M and M2W models separately. This makes the steps to perform adaption more straightforward than those for conventional E2E-TTS models that consist of T2M and M2W models.
We observe that the naturalness gap was larger for male speakers than female ones. 
This is because the speaking style of males speakers is more emotional than that of females here, which can be also confirmed from the scores of their ground-truth utterances.

\section{Summary}
This paper introduces ESPnet2-TTS, an E2E-TTS toolkit extending ESPnet-TTS.
ESPnet2-TTS enhances the flexibility, scalability, and portability of ESPnet-TTS and provides new state-of-the-art TTS models and our own extensions of them.
The experimental results demonstrate that our models can achieve TTS performance comparable to the ground-truth in both single-speaker and multi-speaker settings except for adaptation on small data.

In the future, we will work on training with noisy datasets, which include bad quality recordings, background noise, and transcription errors, as well as E2E speech-to-speech translation.


\section{References}
{
\setstretch{0.85}
\printbibliography
}
\end{document}